# Application of artificial neural networks and genetic algorithms for crude fractional distillation process modeling

Łukasz Pater
PKN ORLEN S.A. Płock, Poland,
Faculty of Mathematics and Computer Science, Nicolaus Copernicus University, Toruń, Poland
Email: pater.lukasz@gmail.com

*Abstract*—This work presents the application of the artificial neural networks, trained and structurally optimized by genetic algorithms, for modeling of crude distillation process at PKN ORLEN S.A. refinery. Models for the main fractionator distillation column products were developed using historical data. Quality of the fractions were predicted based on several chosen process variables. The performance of the model was validated using test data. Neural networks used in companion with genetic algorithms proved that they can accurately predict fractions quality shifts, reproducing the results of the standard laboratory analysis. Simple knowledge extraction method from neural network model built was also performed. Genetic algorithms can be successfully utilized in efficient training of large neural networks and finding their optimal structures.

I. Introduction

Artificial neural networks (ANN) as well as genetic algorithms (GA) are popular machine learning technologies. They were both designed in analogy to structures and processes occurring in nature. Due to their desired properties, they are widely applied to solve various problems of modeling and optimization [1-5]. Although those techniques are utilized mostly separately, together they can extend the range of their possible applications. They can be applied to problems, where it is difficult to recover a clear analytical solution [6,7]. The example can be crude fractional distillation, which is a highly nonlinear process [8]. It also includes many random disturbances caused by many unpredictable factors like mechanical or measurement problems [9]. Because of the above obstacles, data from such process can be perfectly suited to verify the behaviour and properties of artificial neural networks trained and optimized by genetic algorithms.

The aim of this work is to apply artificial neural networks, trained and structurally optimized by the genetic algorithm, to model laboratory quality measurements of main crude oil fractional distillation products.

*A. Artificial neural networks*

Artificial neural network is a structure built by many interconnected basic elements called artificial neurons. It resembles the natural tissues in brain, which consists from many nervous cells. Observation of the natural neurons behaviour resulted in uncovering its basic operation principles and interesting properties. The first mathematical description of artificial neuron was done by McCulloch and Pitts in 1943. Rosenblatt in 1957 refined his idea creating modern artificial neuron called simple perceptron [10,11].

Perceptron artificial neuron can be perceived as a transducer of many input signals giving one output. This led to deriving mathematical description of the neuron, which is presented on Fig. 1 and characterized by equation (1).

$$y = F\left(\sum_{i=0}^{n}(u_i w_i) + w_0\right) \quad (1)$$

where:

y – value of the output signal,
F – activation function,
n – number of input signals,
u – value of input signal number i,
w – weight value of the connection number i.

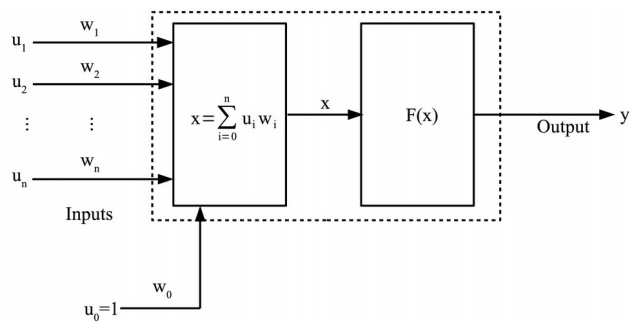

Fig 1. Artificial neuron – simple perceptron [11].

Weights located on input connections are coefficients, which are set during learning process. They can resemble storage of gained knowledge. Every input value is multiplied by weight coefficient and added at the end. This sum is then an argument for activation function of choice. Activation function determines the properties of the artificial neuron. Typically the unipolar sigmoid activation function, defined by equation (2), is commonly used [10, 11]. Chart showing its curve was presented on Fig. 2. Results of this function are included in range from 0 to 1. β parameter in this equation is responsible for the sigmoid function steepness. Learning process is basically equivalent to shape modification of the activation function. The result of this function is the output value, which can be final or become input for another artificial neuron [10,11].

$$F(x) = \frac{1}{1+e^{-\beta x}} \quad (2)$$

where:
F – type of activation function,
x – sum of weights multiplied by input values,
β – sigmoid function steepness parameter.

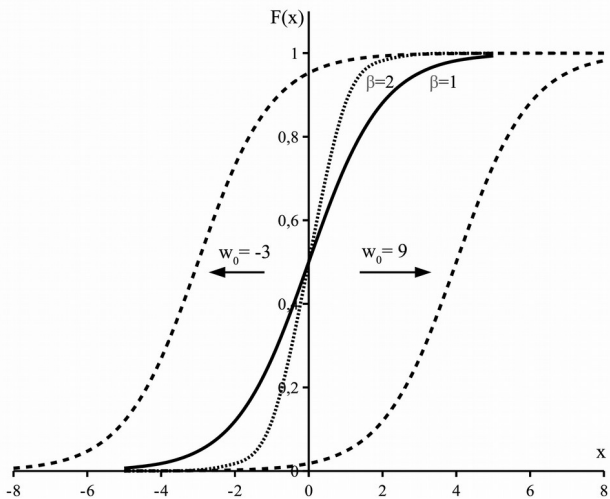

Fig 2. Sigmoid activation function [11].

In order to extend the single perceptron classification capabilities of nonlinear relationships, many artificial neurons are grouped into many layers, creating multi layer perceptrons (MLP). Layers between the input and output layer are called hidden layers. This class of artificial neural network is widely used for a great majority of problems, because it can model highly nonlinear data relationships. Structure of artificial neural network also has to be chosen carefully. Too large structure, consisting of many layers or neurons, may result in overfitting. It means that artificial neural network can classify correctly only training data [10,11]. Choosing the optimal number of layers in multilayer perceptron and number of artificial neurons in hidden layers, can represent a task for the genetic algorithm to solve [12-15].

Artificial neural networks are also not uniform. They can be divided into a diverse group of classes, based on the type of connections between neurons. In this study only multilayer perceptron feedforward networks of fully interconnected neurons are considered, which is used most often. It means that signals they receive are directed from the input to output, without cycles. The principle of artificial neural network operation is based on proper modification of signals that flow through the connections between neurons, by the weights accredited to this connections. The acquisition of knowledge, is performed during neural network learning and it is based on iterative modification of the weights in order to achieve desired output [10, 11].

In this work only supervised type of learning is used. This type of training is based on presenting to the artificial neural network example pairs of input values with corresponding outputs. For each pair difference between expected example output and resulting output is calculated. The overall performance of artificial neural network is measured by error function, which can be defined as root mean square (3). Total training error after one iteration of training is sum of training data sample errors (4) [10, 11].

$$E_s = \frac{\sum_{k=0}^{y}(t_k s_k)^2}{y} \quad (3)$$

$$E_{total} = \sum_{s=0}^{n} E_s \quad (4)$$

where:
$E_s$ – root mean square of two sets t and s,
y – number of outputs,
t – set of model output values,
s – set of calculated output values,
$E_{total}$ – total error of single iteration of training,
n – total number of training data samples.

After that the weights of artificial neural network are updated by applied training algorithm. New result of error function is calculated. One iteration is called epoch. This cycle ends, when the calculated error is lower than arbitrarily set desired value [10, 11].

There are several training algorithms for feedforward multilayer perceptrons, which vary in performance. Typically, the back-propagation (BP) training algorithm is used, because of its speed. It works by minimization of the training error function. In every iteration it follows only those weights values, which gives lowest training error, against the gradient of error function [10, 11]. Unfortunately, most of the gradient-based algorithms, like BP, share the same drawback. While adjusting weights they can easily get stuck in the local minima of the error function, what causes training to stop prematurely, without reaching desired minimal target error [16]. It is narrowing the space of possible solutions. Artificial neural network created in this manner is not able to fully model the training data, showing poor accuracy. There were many attempts to solve this problem by introducing various modifications into the training process itself [11]. One of the different solutions is to utilize genetic algorithm for training [17-23].

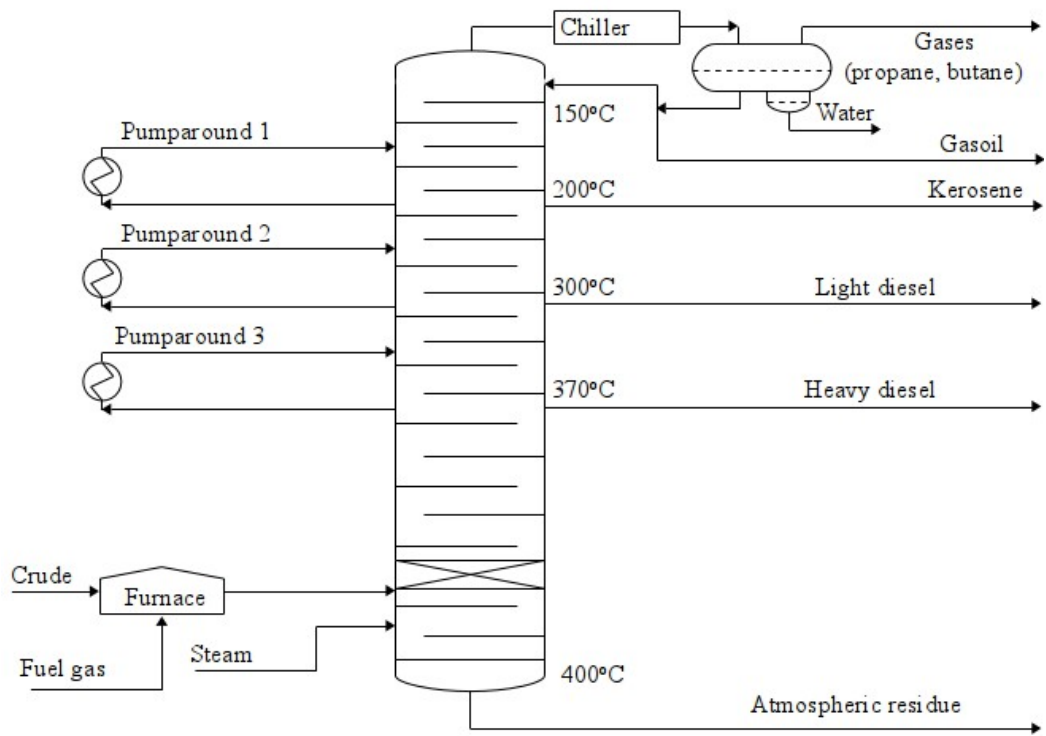

Fig 3. Atmospheric crude distillation column [23].

*B. Genetic algorithm*

Genetic algorithm is a general purpose global random search algorithm. It is mostly used for the large variety of optimization problems. It can also be used for training of artificial neural network. Because it is not using gradient of error function to reach best solution, it is also not sensitive to local minima problem [15].

Design of the algorithm was inspired by observation of natural Darwinian evolution process and survival of the fittest principle. Algorithm is performing operations on the chromosomes, which could be defined as set of features. Population consists of many chromosomes. Establishing a proper measurement of fitness is also crucial in setting correct direction of optimization. Every genetic algorithm performs several operations including [15, 17]:
1. random initialization of the preliminary population,
2. in-loop evaluation of every chromosome by measuring its fitness,
3. comparison with the minimal desired fitness,
4. selection of the fittest subset of chromosomes,
5. perform crossing-over, which is exchange of features from the selected subset of chromosomes,
6. introduce mutations, which are random changes applied to randomly chosen features of the chromosomes,
7. return to the 2nd point.

Genetic algorithm can be used in any optimization problem. It can be used in artificial neural network training or determining its best performing structure. During training, every chromosome for genetic algorithm comprises from all the connection weights from artificial neural network. When searching for best structure of artificial neurons in the network, chromosomes possesses information mostly about number of layers and neurons in each layer.

*C. Crude oil fractional distillation*

Crude oil is a complex mixture of hydrocarbons, that vary in their molecular weight and properties. Due to the above, they also have various boiling temperatures. The main objective of distillation is to separate crude oil into fractions, using those differences [23].

First step of this process is performed in the distillation columns, under atmospheric pressure, in every refinery. Raw crude oil, is pumped directly from the extraction site. Then, it is heated in furnace and enters the atmospheric distillation column. In every column there are a several dozen of trays, which improve the separation by temporarily keeping some amount of liquid. Every column of this type also possesses steam injection and several pumparounds, which are responsible for recovering excessive amount of heat and increasing evaporation rate of hydrocarbons, what further improves separation. Final products are recovered via side draws. The residue from the bottom is directed for vacuum distillation. Efficiency of the distillation depends on side products draw maximization and minimization of the leftover residue. Diagram of crude oil atmospheric distillation column is presented on the Fig. 3 [23].

Every refinery imposes its own quality standards, that distillation process operators have to keep. Quality of each fraction can be changed by manipulation of those key parameters: temperatures or flows of draws, pumparounds. Other important parameters are steam to column, amount of feed to the column and temperature after the furnace. Quality of each fraction is determined by daily laboratory analysis, according to ASTM D86 international standard [8]. It is done by measuring temperature, in which known volume of sample evaporates. Operators adjust the distillation process parameters depending on those analyses [8]. Every laboratory sample analysis is also afflicted with a measurement error, in this case it can be 3-4 degrees.

Although it is possible to predict fraction quality from the temperature measurement based on the chosen column tray using statistical linear regression model, there are many other factors involved, that lowers the accuracy [8]. Distillation is a nonlinear process. Raw data from measurements often contain noise, false values or have a great variability. In order to improve prediction of the fraction quality, we can take into account more process variables. Establishing a clear relation between them is not often viable. Properties of artificial neural networks and genetic algorithm combined, makes them suitable for solving such a problem, avoiding overfitting or local minima problem during training [8, 24-26].

## II. METHODS

### A. Software design

In order to maintain both high performance of artificial neural network training procedure and speed of development, all of the calculation intense programs were written in pure C programming language. User interface, genetic algorithm structure optimization as well as data preparation scripts were written using PHP. Only feedforward multilayer perceptron neural networks were considered in this study.
The task of training algorithm is to find the optimal weight value set, promoting variables of higher model importance. The goal of genetic algorithm structure optimization is to find the neural network configuration with the lowest training error.

### B. Data preparation

Input variables for the artificial neural network model were chosen arbitrarily. Every model created was built using 13 main distillation parameters, which included:
- flows of: product draws, pumparounds, total feed to the column, steam to column;
- temperatures of: crude oil after passing furnace, pumparounds, draw tray, column top;
- pressure in the column.

Through the courtesy of PKN ORLEN S.A. refinery (Płock, Poland), 1 year data readings from mayor parameters of one of the crude oil atmospheric distillation column was extracted, with the sampling frequency of 1 minute. Corresponding laboratory analyses of the fractions quality were also collected.

Data set was then divided equally into two subsets: training and validation data. The main purpose of validation data is to evaluate the ability of the trained artificial neural network to model values of laboratory analyses of fraction samples. It is a simple solution to detect problems with artificial neural network models like overfitting effect or premature ending of a training procedure. Ideal artificial neural network should generalize its knowledge, that means it can model with high accuracy not only training data, but also validation data.

In order to fit the sigmoid activation function, all of the input data values were scaled into 0 and 1 range. Procedure of normalization was performed using equation (5).

$$N(x) = \frac{x - \min(x)}{\max(x) - \min(x)} \quad (5)$$

where $N(x)$ is a result of x normalization.

### C. Structure optimization

Genetic algorithm was applied in feedforward multilayer perceptron artificial neural network structure optimization application. Optimized parameters, that were included in chromosomes, consisted of not only number of artificial neurons, hidden layers but also parameters for training: population size, maximal learning step size, percentage of the fittest chromosomes for crossing-over, number of random mutations per chromosome and crossing-over intensity per chromosome. For every combination of the above parameters, adequate artificial neural network was built and trained. Population number of 10, yielded models with different training errors. Training of a single artificial neural network was limited to 10000 iterations (epochs). Only 5 of the fittest were chosen for creation of a next generation chromosomes, which underwent mutation and crossing-over operations. This process was repeated over 20 generations. The objective of structure optimization, was to determine such parameter combination, that resulted in artificial neural network with lowest training error.

### D. Training process

Every artificial neural network training was performed using training application implementing genetic algorithm. Chromosome was built from all of the weights written in list. Learning parameters and neural network structure was received from the structure optimization application. Training ended when the artificial neural network error converged to the minimal value. Trained weights as well as learning parameters and structure information was written to single text file. Models created in this way, were then utilized to generate output values from the training and validation input data sets.

## III. RESULTS

Quality of the mayor crude distillation products in kerosene, light and heavy diesel fraction, according to international ASTM D86 standard, were determined by artificial neural networks, trained and optimized by genetic algorithm. Temperatures, in which 5, 10, 50, 90, 95, 100 percentage of a sample volume evaporates, was calculated for each fraction product. The results shown similar performance in reproducing laboratory analyses on validation data, so only neural network model for 10% kerosene sample evaporation temperature was discussed as an example.

### A. Genetic algorithm structure optimization

Results of the genetic algorithm structure as well as training parameters optimization are presented in the Table 1. Genetic algorithm, moving within available range set, created total of 200 different neural network models were tested with root mean square (3) training error ranging from 0,0032 to 0,053. Those optimized parameters yielded the fittest artificial neural network with training error of 0,0032.

### B. Learning curve

Learning curve shape visualizes the progress of the training process. It can be helpful in determining the phase of learning and distance to the end. The learning curve of the best artificial neural network is presented on the Fig 4.

### C. Genetic algorithm training

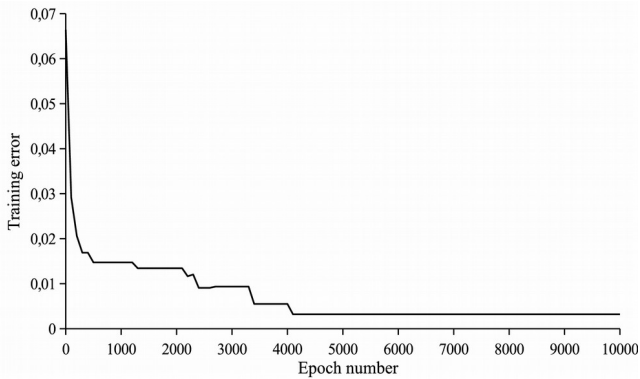

Fig 4. Learning curve of the neural network model for temperature of 10% kerosene sample evaporation.

Fig. 5. presents the performance of generated artificial neural network to model the results of laboratory kerosene sample analyses. For this model only training data set was used.

### D. Model validation

In order to identify the generalization ability of the created model, test data set was used, which was never used in training. Plot shown at Fig. 6. presents the performance of the artificial neural network to mimic the laboratory analyses. It can show how the model is behaving using data previously unseen.

## IV. DISCUSSION

Structure chosen by genetic algorithm of the neural network is a multilayer perceptron with one hidden layer, consisting of 5 neurons. Such configuration is simple in comparison with the number of input values and may hold a promise to avoid over-fitting, This kind of structure is mostly used because of its ability to model nonlinear data. Two other options, which are simple perceptron and two hidden layer perceptron may have not enough potential to either model training data or unnecessarily prolong the training process. In Table 1 we can see that learning parameters were set mostly between their limits. Among training parameters we can also observe a high percentage of killed chromosomes on epochs, high crossing-over rate and population number. This may suggest a great evolutionary pressure for neural network to develop minimal training error.

Learning curve on Fig 4. is of typical shape of every training algorithm. It shows a rapid error function descent with every epoch in the first stage. Then, the curve of the error slows down, because finding better solutions is becoming much harder. After that learning curve reaches plateau, when the training algorithm can not find a better weight combination. Training was completed after 5000 iterations.

TABLE I.
PARAMETERS FOR GENETIC ALGORITHM ARTIFICIAL NEURAL NETWORK STRUCTURE OPTIMIZATION FOR TEMPERATURE OF 10% VOLUME KEROSENE SAMPLE EVAPORATION.

|  | Available range | Result |
|---|---|---|
| Artificial neural network structural parameters | | |
| Number of inputs | - | 13 |
| Number of outputs | - | 1 |
| Number of hidden layers | 0-2 | 1 |
| Number of neurons in 1st hidden layer | 1-10 | 5 |
| Number of neurons in 2nd hidden layer | 1-10 | - |
| Genetic algorithm artificial neural network training parameters | | |
| Population number | 5-60 | 40 |
| Learning step size | 0,05-0,6 | 0,35 |
| Number of mutations per single chromosome | 10-70 | 33 |
| Number of chromosome crossing-over operations | 10-70 | 40 |
| Maximal length of chromosome mutation | 1-10 | 4 |
| Percentage of killed chromosomes on epoch | 0,2-0,8 | 0,6 |
| Sigmoid function steepness parameter | 0,3-0,5 | 0,4 |

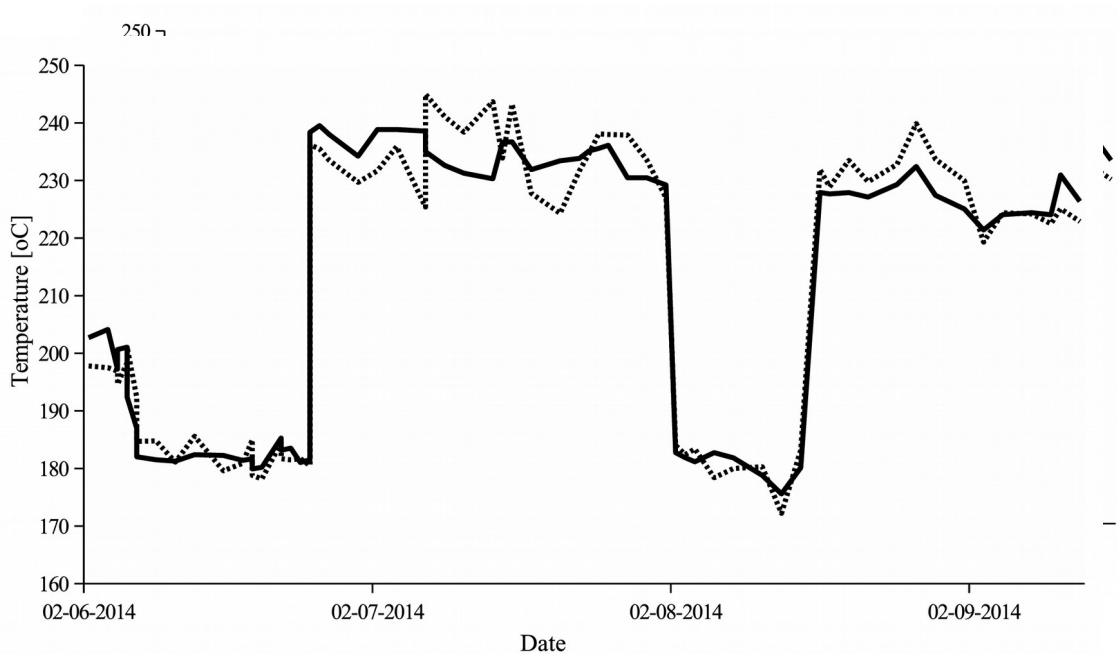

Fig 5. Performance of the artificial neural network to model the temperature of 10% kerosene sample evaporation determined by laboratory analyses during training.

Fig 5. is presenting ability of created artificial neural network to model the training data. We can observe correctly modeled three major changes in the distillation process with a slight deviations on 14.07.2014 and 1.09.2014 The most important chart is presented at Fig 6., when the artificial neural network was calculating quality output based only on the knowledge it retained. We can notice that two major kerosene fraction quality shifts were predicted correctly: 6.03.2014 and 7.05.2004. There were several minor quality changes 13.01.2014, 21.03.2014 and 4.04.2014, which were also foreseen by the artificial neural network model. There were also two cases where the difference between prediction and laboratory analysis is of 4-8 degrees: 10.02.2014 and 12.05.2014. In general model can accurately predict most of the variability of the kerosene fraction quality and can be used as a valuable simulation tool.

The results could have been influenced by many factors. The biggest impact on the models have the data used for training and validation. The data sets used were obtained from a high throughput, industrial crude distillation process under normal operation, so it may contain natural errors like false or low accuracy measurements. Also the behaviour of the plant changes in time. For example, the performance of heat exchangers installed on the pumparounds is decreasing in time due to fouling and they need to be cleaned periodically. Flow measurements on valves also need to be properly calibrated. Another source of possible errors may represent the kerosene fraction sample collection of frequency 2-3 times a day, which might not be collected on the exact time labelled or properly handled. Additionally, the laboratory analysis is also naturally affected with error.

However, the results obtained from the artificial neural network model are well enough to quickly provide valuable process information, without the need to wait for the laboratory analysis results. It may be a on-line soft sensor for kerosene product quality.

Artificial neural network model is often considered as a black box model, because it is not possible to recover a clear relationship from the weights between input and output [27]. However, there are ways to recover basic rules from trained neural network knowledge, explaining the output result [28].

One of the simple method of approximate knowledge extraction form trained artificial neural network, is casual index technique, suggested in [29], which is described by equation (6) [29].

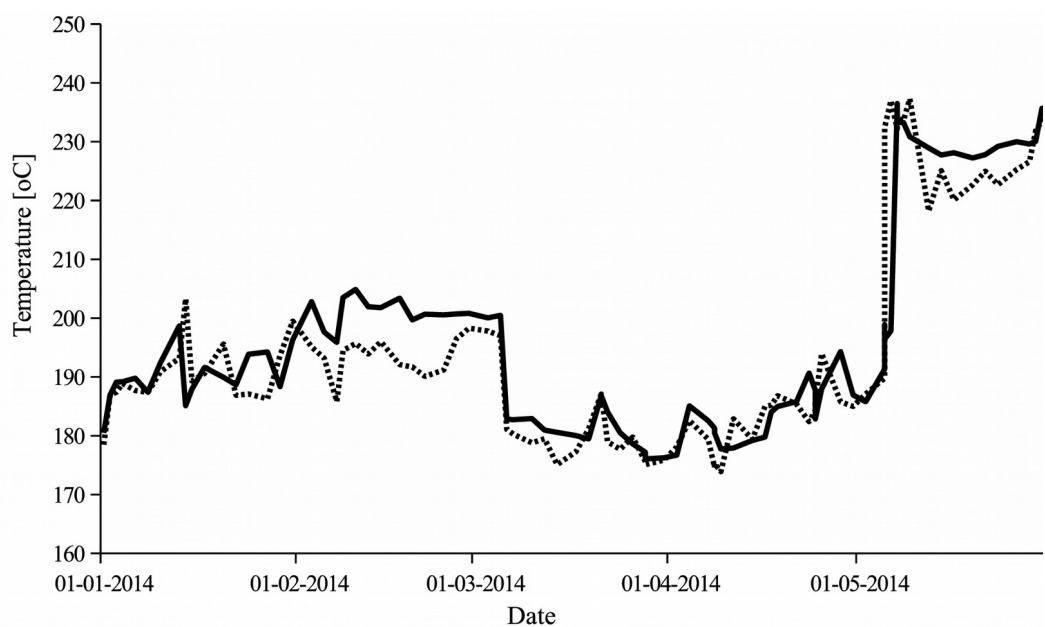

Fig 6. Performance of the artificial neural network to model the temperature of 10% kerosene sample evaporation determined by laboratory analyses during testing.

$$CI = \sum_{j=1}^{h} \left( w_{yj} w_{yi} \right) \quad (6)$$

where:
- h – total number of hidden neurons,
- j – single hidden neuron,
- y – output,
- $w_{yj}$ – weights from output to hidden neurons,
- $w_{ju}$ – weights from hidden neurons to input.

It measures the estimate impact magnitude of every input value on the single output. Casual index of a single input can be perceived as sum of multiplied weights that lay on paths from output to that input [29]. Inputs with a positive CI values are most significant, because they mostly influence output [30].

Calculated CI values for the artificial neural network in this work are presented in the Table II.

TABLE II.
CASUAL INDEX VALUES FOR THE ARTIFICIAL NEURAL NETWORK TRAINED FOR KEROSENE QUALITY. THE MOST SIGNIFICANT INPUTS IN BOLD.

| CI | Input variable description |
|---|---|
| 141 | Kerosene draw |
| 76 | Kerosene pumparound |
| 61 | Steam to column |
| 41 | Steam to side stripper |
| 37 | Top column temperature |
| 15 | Total feed |
| -8 | Kerosene tray temperature |
| -9 | Furnace temperature |
| -35 | Column top pressure |
| -43 | Top reflux |
| -56 | Light diesel pumparound |
| -56 | Kerosene pumparound temperature |
| -151 | Light diesel draw |

Casual index method presented that neural network correctly identified mayor quality parameter, which is kerosene draw. Another important parameters are feed and top temperatures [8]. Kerosene pumparound and steam amount are also important factors responsible for separation in the kerosene section of distillation column. Based on this we can can conclude that neural network model trained by genetic algorithm uses valid parameters.

In order to ensure the correctness of the output values from the trained neural network, it is vital to use input values in the same range when training. Using input values, outside the range of training may lead to false results [31],

One of the main advantage of artificial neural network model trained and optimized by genetic algorithm is simplicity of use. Model generation process, using written software, is fully automated and requires very little maintenance. Automatic model update can be used after certain period of time or on demand, in order to fit a new operation conditions of the modeled subject, like a distillation process in a refinery.

It is also easy to notice that genetic algorithm can efficiently train even large neural networks, given its learning parameters are set properly. Using only basic means of natural selection, it took 5000 genetic algorithm training iterations to set sum of 76 weights of the best performing artificial neural network (Table 1). That also involved assessing the performance of 40 chromosomes in population, 76 weights each. Addition of another hidden layer or increasing number of artificial neurons can double that number. Still, genetic algorithm can successfully train a valid neural network model, even working on such a long chromosome [22].

Future work involves using created artificial neural network and genetic algorithm modeling software for building simulator of the whole industrial process unit such as crude distillation units. This would enable to test the behaviour of the plant and identify its performance boundaries as well as serve for people training purposes. This research can be also continued by performing a comparison between the most popular neural networks training algorithm back-propagation and genetic algorithm in terms of speed and efficiency of training.

V. CONCLUSIONS

Artificial neural networks, which weights and structure has been optimized by genetic algorithm, proved that they can be successfully utilized for modeling complex relationships using raw process data. Genetic algorithm can operate even on a large parameter chromosomes, given the amount of crossing-over and mutation operations are set correspondingly. Presented techniques have a great potential not only in the crude oil refining industry, but also in general model prediction as well as global optimization problems.

REFERENCES

[1] Smith K. A., Gupta J. N. D., *Neural networks in business: techniques and applications for the operations researcher*, Computers & Operations Research 27, 2000, 1023-1044.
[2] Wu R. C., *Neural network models: Foundations and applications to an audit decision problem*, Annals of Operations Research 75, 1997, 291-301.
[3] Karul C., Soyupak S., Yurteri C., *Neural network models as a management tool in lakes,* Hydrobiology 408/409, 1999, 139-144.
[4] Yin-ju B., *Application of genetic BP network to discriminating earthquakes and explosions*, Acta Seismologica Sinica, 2002, 540-549.
[5] Krongrad A., Lai S., *Artificial neural networks in urology: Con,* Urology 54, 1999, 949-951.
[6] Dokur Z., *Segmentation of MR and CT Images Using a Hybrid Neural Network Trained by Genetic Algorithm*, Neural Processing Letters 16, 2002, 211-225.
[7] Henderson C. E., Potter W. D., *Predicting Aflatoxin Contamination in Peanuts:A Genetic Algorithm/Neural Network Approach,* Applied Intelligence 12, 2000, 183-192.
[8] Brambilla A., *Distillation control and optimization*, McGraw-Hill Education, 2014, 309-319.
[9] Kaes G. L., *Some Practical Aspects Of Modeling Crude Oil Distillation*, VMGSimUser'sManual, 2012, 1-18.
[10] Rojas R., *Neural Networks A Systematic Introduction*, Springer-Verlag, Berlin, 1996, 1-28, 57-77, 429-449.
[11] Kriesel D., *A Brief introduction to Neural Networks,* www.dkriesel.com, 2007, 15-37.
[12] Stastny J., Skorpil V., *Genetic Algorithm and Neural Network,* Proceedings of the 7[th] WSEAS International Conference on Applied Informatics and Communications, Athens, 2007, 345-349.

[13] Periaux J., Winter G., *Genetic Algorithms in Engineering and Computer Science*, John Wiley& Sons, 1995, 1-15.
[14] Fiszelew A., Britos P., Ochoa A., Merlino H., Fernández E., García-Martínez R., *Finding Optimal Neural Network Architecture Using Genetic Algorithm,* Research in Computer Science 27, 2007, 15-24.
[15] Jiang J., *BP Neural Network Algorithm Optimized by Genetic Algorithm and Its Simulation*, International Journal of Computer Science Issues 10, 2013, 516-519.
[16] Joy C. U., *Comparing the Performance of Backpropagation Algorithm and Genetic Algorithms in Pattern Recognition Problems*, International Journal of Computer Information Systems 2, 2011, 7-12.
[17] Montana D. J., Davis L., *Training Feedforward Neural Networks Using Genetic Algorithms*, Proceedings of the 11th international joint conference on Artificial Intelligence, 1989, 762-767.
[18] Vishwakarma D. D., *Genetic Algorithm based Weights Optimization of Artificial Neural Network*, International Journal of Advanced Research in Electrical, Electronics and Instrumentation Engineering 1, 2012, 206-211.
[19] Montana D. J., *Neural Network Weight Selection Using Genetic Algorithms*, Intelligent Hybrid Systems, 1995, 85-104.
[20] Sexton R. S., Dorsey R. E., Johnson J. D., *Toward global optimization of neural networks: a comparison of the genetic algorithm and backpropagation*, Decision Support Systems 22, 1998, 171-185.
[21] Sexton R. S., Dorsey R. E., *Reliable Classification Using Neural Networks: A Genetic Algorithm and Backpropagation Comparison,* Decision Support Systems 30, 2000, 11-22.
[22] Korning P. G., *Training of Neural Networks by means of Genetic Algorithms Working on very long Chromosomes*, Journal of Neural Systems 6, 1995, 299-316.
[23] Fahim M. A., Al.-Sahhaf T. A., Elkilani A. S., *Fundamentals of Petroleum Refining*, Elsevier B. V., 2010, 69-80.
[24] Quin S. J., *Neural Networks for Intelligent Sensors and Control - Practical Issues and Some Solutions,* 1996, 215-236.
[25] Michalopoulos J., Papadokonstadakis S., Arampatzis G., Lygeros A., *Modeling of an Industrial Catalytic Cracking Unit Using Neural Networks*, Trans IChemE 79, 2001, 137-142.
[26] Bo C. M., Li J., Zhang S., Sun C. Y., Wang Y. R., *The Application of Neural Network Soft Sensor Technology to an Advanced Control System of Distillation Operation*, Proceedings of the International Joint Conference on Neural Networks, 2003, 1054-1058.
[27] Smith K. A., Gupta J. N. D., *Neural networks in business: techniques and applications for the operations researcher*, Computers & Operations Research 27 2000, 1023-1044.
[28] Feraud E., Clerot F., *A methodology to explain neural network classification*, Neural networks 15, 2002, 237-246.
[29] Boger Z., Guterman H., *Knowledge Extraction from Artificial Neural Networks Models,* IEEE International Conference On Systems, Man, And Cybernetics 4, 1997, 3030-3035.
[30] Pérez-Magariño S., Ortega-Heras M., González-San José M.L., Boger Z., *Comparative study of artificial neural network and multivariate methods to classify Spanish DO rose wines*, Talanta 62, 2004, 983–990.
[31] Kurd Z., Kelly T., Austin J., *Developing artificial neural networks for safety critical systems*, Neural Computing & Applications 16, 2007, 11-19.